\documentclass{article}

\usepackage{silence}
\usepackage{styles/LL}

\usepackage{enumitem}

\usepackage[normalem]{ulem}

\usepackage{graphicx}
\usepackage{amsmath}
\usepackage{dsfont}
\usepackage{amssymb}

\usepackage{float}

\usepackage{siunitx}
\usepackage{tikz}
\usepackage{pgfplots}
\usetikzlibrary{positioning, fit, shapes.geometric}
\pgfplotsset{compat=1.14}

\usepackage{amsthm}

\usepackage{import}
\usepackage{standalone}

\usepackage{mathabx} %

\usepackage{bm}

\usepackage{comment}

     \usepackage[nonatbib,preprint]{neurips_2020}

\usepackage[utf8]{inputenc} %
\usepackage[T1]{fontenc}    %
\usepackage{hyperref}       %
\usepackage{url}            %
\usepackage{booktabs}       %
\usepackage{amsfonts}       %
\usepackage{nicefrac}       %
\usepackage{microtype}      %

\usepackage{cleveref}

\title{Ensembles of Generative Adversarial Networks \\ for Disconnected Data}

\author{%
    Lorenzo Luzi \\
    Rice University\\
    \texttt{lorenzo.luzi@rice.edu} \\
    \And
    Randall Balestriero\\
    Rice University \\
    \texttt{randallbalestriero@gmail.com} \\
    \And
    Richard G. Baraniuk \\
    Rice University \\
    \texttt{richb@rice.edu} \\
}

\tikzset{%
    font=\small
}

\pgfplotsset{
    title style={yshift=-2.0ex}
}
\pgfplotsset{
    x label style={yshift=1.0ex}
}

\begin{document}

\maketitle

\begin{abstract}

Most current computer vision datasets are composed of disconnected sets, such as images from different classes. We prove that distributions of this type of data cannot be represented with a continuous generative network without error. They can be represented in two ways: With an ensemble of networks or with a single network with truncated latent space. We show that ensembles are more desirable than truncated distributions in practice. We construct a regularized optimization problem that establishes the relationship between a single continuous GAN, an ensemble of GANs, conditional GANs, and Gaussian Mixture GANs. This regularization can be computed efficiently, and we show empirically that our framework has a performance sweet spot which can be found with hyperparameter tuning. This ensemble framework allows better performance than a single continuous GAN or cGAN while maintaining fewer total parameters. 
\end{abstract}

\section{Introduction}

Generative networks, such as generative adversarial networks (GANs)~\cite{goodfellow2014generative} and variational autoencoders~\cite{kingma2013auto}, have shown impressive performance  generating highly realistic images 
that are not observed in the training set~\cite{biggan,karras2017progressive,styleGAN,styleGAN2}. 
In this work, we will focus on the effect of disconnected data on GAN performance.
Here disconnected means that we assume that data points are drawn from an underlying topologically disconnected space. As an intuitive example, consider the collection of all images of badgers and all images of zebras. These two sets are disconnected because images of badgers do not look like images of zebras. Others have studied such data~\cite{disconnected,MGAN,MMGAN}.

We prove that one cannot use a single continuous generative network to learn a data distribution perfectly under the disconnected data model, defined below in \cref{sec:disconnectedAssumption}.
Because generative networks are continuous, we cannot map a connected latent space ($\bbR^\ell$) into the disconnected image space, resulting in the generation of data outside of our true data space. In~\cite{disconnected}, the authors empirically study disconnected data but do not formally prove the results in this paper. In addition, they use a completely unsupervised approach and try to find the disconnected components as part of learning. We use class labels and hence work in the supervised learning regime.

Our solution to tackling disconnected data is to study ensembles of GANs. 
Ensembles of GANs are not new, e.g., see~\cite{dualDiscriminator,MADGAN,tolstikhin2017adagan,MIXGAN}, but there is limited theoretic study of them to our best knowledge.
We show that ensembles are optimal for disconnected data and study their relation to single GANs. Specifically, we design a theoretic framework that demonstrates how a single GAN, an ensemble of GANs, a conditional GAN, and a Gaussian mixture GAN are all related, something that we have not seen in ensemble GAN literature. This leads us to consider ensembles that are regularized to encourage parameter sharing.

We primarily focus on theory in this paper, however we conduct experiments to show that the performance of our models increases when we use an ensemble of WGANs over a single WGAN on the CIFAR-10 dataset~\cite{cifar10}.
We evaluate performance using FID~\cite{FID}, MSE to the training set~\cite{metz2016unrolled}, Precision, and Recall~\cite{PR}. 
This performance increase can be decomposed into three contributing factors: 1) the ensemble has more parameters and hence is more overparameterized, 2) the ensemble better captures the structure of the data, and 3) parameter sharing among ensemble networks allows successful joint learning.

We summarize our contributions that apply to generative networks in general as follows: 
\begin{itemize}[leftmargin=0.5cm]

    \item We show that generative networks, which are continuous functions, 
    cannot learn the data distribution if our data is disconnected (\cref{sec:continuousNetworksCantLearn}). This disconnected data model is discussed in \cref{sec:disconnectedAssumption} and we argue that it is satisfied in very common datasets, such as MNIST~\cite{mnist}, CIFAR-10~\cite{cifar10}, and ILSVRC2012~\cite{ILSVRC15}. Restricting the generator to a disconnected subset of the domain is one solution (which \cref{sec:restrictingG}), but we study a better solution: using ensembles.
    
    \item 
    We first show how single GANs and ensembles are related (\cref{sec:ensemblesIntro}). We then prove that ensembles are optimal under our data model (\cref{sec:ensemblesAreOptimal}). Finally, we demonstrate that there is an equivalence between ensembles of GANs and common architectures such as cGANs and GM-GANs due to parameter sharing between ensemble components (\cref{sec:GANsThatAreSecretlyEnsembles}).
    
    \item We show that an ensemble of GANs performs better than a single GAN (\cref{sec:exp_full_ensemble}). This is true even if we reduce the parameters used in an ensemble so that it has less total parameters than a single network (\cref{sec:exp_equi_ensemble}). Finally, we show that parameter sharing among ensemble networks leads to better performance than a single GAN (\cref{sec:exp_L1_ensembles}) or even a cGAN (\cref{sec:cGANs_suck}).
\end{itemize}

\section{Background and Related Work}

\subsection{Generative adversarial networks (GANs)}
\label{sec:backgroundGANs}
GANs are generative neural networks that use an adversarial loss, typically from another neural network. In other words, a GAN consists of two neural networks that compete against each other. The generator $G: \bbR^\ell \to \bbR^p$ is a neural network that generates $p$-dimensional images from an $\ell$-dimensional latent space. The discriminator $D: \bbR^p \to (0,1)$ is a neural network which is trained to classify between the training set and generated images. Both $G$ and $D$ are continuous functions. 

$G$ has parameters $\bm \theta_G \in \bbR^{|\bm \theta_G |}$, where $|\bm \theta_G |$ is the possibly infinite cardinality of $\bm \theta_G$. Similarly, $D$ has parameters $\bm \theta_D \in \bbR^{|\bm \theta_D|}$. The latent, generated, and data distributions are $P_\vz, P_G$, and $P_{\mcX}$, respectively. We train this network by solving the following optimization problem:
\begin{align}
    \min_{\bm \theta_G} \max_{\bm \theta_D} V(\bm \theta_G, \bm \theta_D) 
    = \min_{\bm \theta_G} \max_{\bm \theta_D} \bbE_{\vx \sim P_\mcX}
    [\log D( \vx)] 
    + 
    \bbE_{\vz \sim P_\vz} [\log (1 - D(G(\vz)))].
    \label{eq:basic_loss}
\end{align}
Here we write $\min$ and $\max$ instead of $\minimize$ and $\maximize$ for notational compactness, but we are referring to an optimization problem.
The objective of this optimization is to learn the true data distribution, i.e. $P_G = P_\mcX$. Alternatively, we can use the Wasserstein distance instead of the typical cross entropy loss: $V(\bm \theta_G, \bm \theta_D) = \bbE_{\vx \sim P_\mcX}D(\vx) - \bbE_{\vz \sim P_\vz}D(G(\vz))$ 
restricted to those $\bm \theta_G, \bm \theta_D$ which force $D$ to be $1$-Lipschitz as done in the WGAN paper~\cite{arjovsky2017wasserstein}. 
Thus, we will use function $V: \bbR^{|\bm \theta_G| \times |\bm \theta_D|} \to \bbR$ to denote a general objective function that we want to optimize, which can potentially be restricted to a subset of the original domain.

\subsection{GANs that treat subsets of data differently}
\label{sec:GANsTreatingClassesDifferent}

\paragraph{Ensembles.}
Datasets with many different classes, such as ILSVRC2012~\cite{ILSVRC15}, are harder to learn in part because the relationship between classes is difficult to quantify. Some models, such as AC-GANs~\cite{AC-GANS}, tackle this complexity by training different models on different classes of data in a supervised fashion. In the AC-GAN paper,  the authors train $100$ GANs on the $1000$ classes of ILSVRC2012. The need for these ensembles is not theoretically studied or justified beyond their intuitive usefulness.

Several ensembles of GANs have been studied in the unsupervised setting, where the modes or disconnected subsets of the latent space are typically learned~\cite{MMGAN,MGAN, disconnected} with some information theoretic regularization as done in~\cite{infoGAN}. These are unsupervised approaches which we do not study in this paper. Models such as SGAN~\cite{SGAN} and standard GAN ensembles~\cite{wang2016ensembles} use several GANs in part to increase the capacity or expressibility of GANs. Other ensembles, such as Dropout-GAN~\cite{mordido2018dropout}, help increase robustness of the generative network.%

\paragraph{Conditional GANs (cGANs).}
Conditional GANs~\cite{cgans} attempt to solve the optimization problem in (\ref{eq:basic_loss}) by conditioning on the class
$\vy$, a one-hot vector. The generator and discriminator both take $\vy$ as an additional input. This conditioning can be implemented by having the latent variable be part of the input. For example, the input to the generator will be $[\vz^T \ \vy^T]^T$ instead of just $\vz$. 

Typically, we see that traditional cGANs have the following architecture modification. The first layer has an additive bias that depends on the class vector $\vy$ and the rest is the same. For example, consider a multilayer perceptron, with matrix $\mW$ in the first layer. Converting this network to be conditional would result in the following modification to the matrix in the first layer:
\[
    \mW_{\text{conditional}} 
    \left[ \begin{matrix}
    \vx \\ \vy
    \end{matrix}\right]
    =
    \left[ \begin{matrix}
    \mW & \mB
    \end{matrix}\right]
    \left[ \begin{matrix}
    \vx \\ \vy
    \end{matrix}\right]
    =
    \mW \vx + \mB \vy
    =
    \mW \vx + \mB_{\cdot, k}.
\]
Hence, we can think of $\mB$ as a matrix with columns $\mB_{\cdot, k}, k \in \{ 1, \dots, K\}$ being bias vectors and $\mW$ being the same as before. We pick a bias vector $\mB_{\cdot, k}$ based on what class we are conditioning on  but the other parameters of the network are held the same, independent of $k$. This is done to both the generator and the discriminator. Some cGANs condition on multiple layers, such as BigGAN~\cite{biggan}.

\paragraph{Gaussian Mixture GANs (GM-GANs).}
The latent distribution $P_\vz$ is typically chosen to be either uniform, isotropic Gaussian, or truncated isotropic Gaussian~\cite{goodfellow2014generative, radford2015unsupervised, biggan}.
We are not restricted to these distributions; research has been conducted in extending and studying the affect of using different distributions, such as a mixture of Gaussians~\cite{gm-gans,DeLiGAN}. The parameters of the Gaussian mixture are learned during training, which makes backpropagation difficult to calculate. This is tackled by using the re-parameterization trick. This trick is implemented by passing an isotropic Gaussian through a linear transformation to get the corresponding mixture Gaussian. Hence, we pick the linear transformation based on the class vector $\vy$. Thus, we have a linear layer $(\mW_k, \vb_k)$ that depends on the class $k \in \{ 1, \dots, K\}$ while the rest of the parameters of the network are the same. Notice that if $\vz \sim \mcN(0, \mI)$, then $\vx  = \mW_k \vz + \vb_k \sim \mcN(\vb_k, \mW_k \mW_k^T)$~\cite{marola1979multivariate}.

\section{Continuous generative networks cannot model distributions drawn from disconnected data}

In this section, we study generative networks in general. This applies to a very general class of models, including GANs and variational autoencoders.

\subsection{Disconnected data model}
\label{sec:disconnectedAssumption}

We begin with a data model that captures data structure which is very commonly satisfied. Almost all datasets with class labels satisfy this model; we provide some examples below.

\begin{definition}[\bf Disconnected data model]
We assume that our data lies on $K$ disjoint, closed sets $\mcX_k \subset \bbR^p, k\in\{ 1, \dots, K\}$ so that the whole data lies on the disjoint union of each component: $\bigcupdot_{k=1}^K \mcX_k = \mcX$.
Moreover, we assume that each component $\mcX_k$ is connected. We then draw data points from these sets in order to construct our finite datasets.
\label{ass:disconnected}
\end{definition}

In~\cref{ass:disconnected}, we let each $\mcX_k$ be closed 
in order to remove the degenerate case of having two components $\mcX_k$ and $\mcX_j$ that are arbitrarily close to one another, which is possible if we only assume that $\mcX$ is disjoint. If that is the case, there are trivial counter-examples (see the supplementary material) to the theorems proved below.
In addition, one could assume that $\mcX$ is a manifold, but it is not necessary for our analysis.

\begin{lemma}
Under \cref{ass:disconnected}, $\mcX$ is a disconnected set.
    \label{lemma:disconnected}
\end{lemma}

Disconnected datasets are ubiquitous in machine learning~\cite{disconnected,MGAN,MMGAN}.
For example, datasets with discrete labels (e.g., typical in classification problems) will often be disconnected.
More specifically, consider the ILSVRC2012 dataset~\cite{ILSVRC15} which consists of 1000 classes of images of real-world objects. 
Two of these classes are badger images  and zebra images. 
Since we have yet to see a badger that looks like a zebra, it is intuitively clear that the set of of all badger images is disjoint from the set of all zebra images.
We study this disconnected data property because generative networks are unable to learn the distribution supported on such a dataset, as shown below in \cref{sec:continuousNetworksCantLearn}.

However, not all machine learning datasets are disconnected.
Some classification problems involve datasets with classes that are not disjoint. 
For example, we argue that with MNIST~\cite{mnist} the $7$ and $1$ class are not disjoint, because some images from these two classes look like each other; however, the rest of the classes are disjoint.
Finally, video datasets, such as~\cite{soomro2012ucf101}, are also not disconnected, because from frame to frame the video does not change much but the action class can change. This implies that different classes can be arbitrarily close to each other and are not disconnected.
In general, datasets do not have to have disjoint classes but many datasets conform to~\cref{ass:disconnected}.
In this paper, we study only disconnected datasets which are known to be disconnected by the use of class labels.

\subsection{Continuous generative networks cannot 
represent a disconnected data distribution exactly}
\label{sec:continuousNetworksCantLearn}
In this section, we demonstrate how, under \cref{ass:disconnected}, continuous generative networks cannot effectively learn the true data distribution. This result independent of any learning algorithm.

Suppose that $(\Omega, \mcF, P_\vz)$ is a probability space with $P_\vz$ being the distribution of the random vector $\vz: \Omega \to \bbR^\ell$. We assume that $P_\vz$ is equivalent to the Lebesgue measure $\lambda$. This just means that $\lambda(\vz(A)) = 0$ if and only if $P_\vz(A) = 0$ for any set $A \in \mcB$. This is true for a Gaussian distribution, for example, which is commonly used as a latent distribution in GANs~\cite{arjovsky2017wasserstein}.
The transformed (via the generative network $G$) random vector $\vx = G \circ \vz: \Omega \to \bbR^p$ is determined by the original probability measure $P_\vz$ but is defined on the induced probability space $(\Omega', \mcF', P_G)$.

\begin{theorem}
The probability of generating samples outside of $\mcX$ is positive: 
$
    P_G (\vx \in \bbR^p \backslash \mcX) > 0.
$
\label{thm:probPositive}
\end{theorem}

The continuity of $G$ is
the fundamental reason why \cref{thm:probPositive} is true. A continuous function cannot map a connected space to a disconnected space.
This means that all generative networks must generate samples outside of the dataset if the data satisfies \cref{ass:disconnected}. 

Suppose that our data is generated from the true random vector $\vx_{\text{data}}: \Omega' \to \bbR^p$ using the probability distribution $P_\mcX$. Also, suppose that we learn  $P_G$ by training a generative network.

\begin{corollary}
Under~\cref{ass:disconnected}, we have that $d(P_G, P_\mcX) > 0$ for any distance metric $d$ and any learned distribution $P_G$. 
\label{cor:distPositive}
\end{corollary}

From~\cref{cor:distPositive} we see that learning the data distribution will incur irreducible error under \cref{ass:disconnected} because our data model and the model that we are trying to train do not match. Hence, we need to change which models we consider when we train in order to best reflect the structure of our data. At first thought a discontinuous $G$ might be considered, but that would require training $G$ without backpropagation. Instead, we focus on restricting $G$ to a discontinuous domain (\cref{sec:restrictingG}) and training an ensemble of GANs (\cref{sec:ensemblesIntro}) as two possible solutions.

\subsection{Restricting the generator to a disconnected subset of the latent distribution}
\label{sec:restrictingG}

In this section, we study how we can remove the irreducible error in \cref{thm:probPositive} from our models after training. Suppose that we train a generator $G$ on some data so that $G(\bbR^{\ell}) \supset \mcX$. Therefore, we can actually generate points from the true data distribution. We know that the distributions cannot be equal because of~\cref{thm:probPositive}, implying that if we restrict the domain of $G$ to the set $Z = G^{-1}(\mcX)$ then $G(Z) = \mcX$.
The next theorem shows how the latent distribution is related to restricting the domain of $G$.

\begin{theorem}[\bf Truncating the latent space reduces error]
\label{thm:truncatingPz}
Suppose that $P_\vz(Z) > 0$ and let the generator $G$ learn a proportionally correct distribution over $\mcX$. In other words, there exists a real number $c \in \bbR$ so that
\[
    P_G(A) = c P_\mcX(A) \quad \quad A \in \mcF', A \subset \mcX.
\]
Then, we use the truncated latent distribution defined by $P_{\vz_T}(B) = 0$ for all $B \in \mcF$ that satisfy $B \cap Z = \emptyset$. This allows us to learn the data distribution exactly, i.e.
\[
    P_{G|_Z}(A) = P_\mcX(A) \quad \quad A \in \mcF'.
\]
\end{theorem}

We write $P_{G|_Z}$ because by truncating the latent distribution, we effectively restrict $G$ to the domain $Z$.~\cref{thm:truncatingPz} shows that if we learn the data distribution approximately by learning a proportional distribution, then we can learn the true data distribution by truncating our latent distribution. 
By 4.22 in~\cite{babyRudin}, $Z$ must be disconnected, which implies that a disconnected latent distribution is a solution to remove the irreducible error in \cref{thm:probPositive}.

Although \cref{thm:truncatingPz} suggests that we truncate the latent distribution, this is not a good idea for several reasons. 
First, the latent distribution cannot be truncated without knowing a closed form expression for $P_G$. 
Second, we may learn the disconnected set $Z$ by training a mixture distribution for $P_\vz$ as is done in~\cite{gm-gans, DeLiGAN}. The problem with this is that the geometric shape of $Z$ is restricted to be spherical or hyperellipsoidal. 
Finally, before truncating the latent space, we need to train a generative network to proportionally learn the data distribution, which is impossible to confirm.

Given these limitations, we introduce the use of ensembles of generative networks in \cref{sec:ensemblesIntro}. This class of models addresses the issues above as follows.
First, we will not need to have access to $P_G$ in any way before or after training. 
Second, knowing the geometric shape of $Z$ is no longer an issue because each network in the ensemble is trained on the connected set $\mcX_k$ instead of the disconnected whole $\mcX$.
Finally, since the $k$-th network will only need to learn the distribution of $\mcX_k$, we reduce the complexity of the learned distribution and do not have to confirm that the distribution learned is proportionally correct.

\section{Ensembles of GANs}
\label{sec:ensemblesOverall}

We demonstrate how to train ensembles of GANs practically and relate ensembles to single GANs and other architectures, such as cGANs. We focus on GANs in this section for concreteness; therefore, we  study an ensemble of discriminators as well as generators.

\subsection{Ensemble of GANs vs.\ a single GAN}
\label{sec:ensemblesIntro}

When we have an ensemble of GANs, we will write $G_k: \bbR^\ell \to \bbR^p$ as the $k$-th generator with parameters $\bm \theta_{G_k} \in \bbR^{|\bm \theta_G|}$ for $k \in \{ 1, \dots, K\}$, where $K$ is the number of ensemble networks.
We assume that each of the generators has the same architecture, hence $|\bm \theta_{G_i}| = |\bm \theta_{G_j}|$ for all $i, j \in \{ 1, \dots, K\}$; thus we drop the subscript and write $|\bm \theta_G|$. Likewise, we write $D_k: \bbR^p \to [0,1]$ for the $k$-th discriminator with parameters $\bm \theta_{D_k} \in \bbR^{|\bm \theta_D|}$ since the discriminators all have the same architecture. The latent distribution is the same for each ensemble network: $P_\vz$. The generated distributions will be denoted $P_{G_k}$. 

For concreteness, we assume here that $K$ is the number of classes in the data; for MNIST, CIFAR-10, and ILSVRC2012, $K$ would be $10$, $10$, and $1000$, respectively. If $K$ is unknown, then an unsupervised approach~\cite{MGAN,disconnected} can be used.

We define a parameter $\bm \pi \in \bbR_+^K$ so that $\sum_{k=1}^K \pi_k = 1$. We then draw a one-hot vector $\vy \sim$ Cat$(\bm \pi)$ randomly and generate a sample using the $k$-th generator if the $k$-th component of $\vy$ is $1$. Hence, we have that a generated sample is given by $\vx = G_k(\vz)$. This ensemble of GANs is trained by solving the following optimization problem:
\begin{equation}
    \min_{\bm \theta_{G_k}} \max_{\bm \theta_{D_k}} V(\bm \theta_{G_k}, \bm \theta_{D_k})
    \label{eq:ensemble_loss}
\end{equation}
for $k \in \{ 1, \dots, K\}$. Note that with an ensemble like this, the overall generated distribution $P_G (\vx) = \sum_{k=1}^K \pi_k P_{G_k}(\vx) $
is a mixture of the the ensemble distributions. This makes comparing a single GAN to an ensemble challenging.

In order to compare a single GAN to ensembles, we define a new hybrid optimization function
\begin{align}
    \min_{ \bm \theta_{G_1}, \dots, \bm \theta_{G_K}}
    \left(
    \max_{\bm \theta_{D_1}, \dots, \bm \theta_{D_K}} 
    \sum_{k=1}^K V(\bm \theta_{G_k}, \bm \theta_{D_k}) 
    \text{ s.t.} \sum_{\substack{k=1 \\ j = k}}^K \|\bm \theta_{D_j} - \bm \theta_{D_k} \|_0 \leq t
    \right)
    \text{ s.t.} \sum_{\substack{k=1 \\ j = k}}^K \|\bm \theta_{G_j} - \bm \theta_{G_k} \|_0 \leq t,
    \label{eq:hybrid_lossL0}
\end{align}
where $\| \cdot \|_0 = 1$ denotes the $\ell_0$ "norm," which counts the number of non-zero values in a vector. Thus, $t \geq 0$ serves as a value indicating how many parameters are the same across different networks, which is more general than having tied weights between networks~\cite{MADGAN}. We penalize the parameters because it is convenient, although it is not equivalent to, penalizing the functions themselves. This is true because $\bm \theta_{G_k} - \bm \theta_{G_j} = 0 \implies G_k - G_j = 0$ but the converse is not true.  We show the behavior of~\cref{eq:hybrid_lossL0} as we vary $t$ in the next theorem.
\begin{theorem}
Let $G$ and $D$ be the generator and discriminator network in a GAN. Suppose that for $k \in \{ 1, \dots, K\}$ we have that $G_k$ and $D_k$ have the same architectures as $G$ and $D$, respectively. Then,
\begin{enumerate}[label=\roman*)]
    \item Suppose that $t \geq \max\Big\{ K\frac{K-1}{2}|\bm \theta_D|, K\frac{K-1}{2}|\bm \theta_G|\Big\}$. Then for all $k \in \{ 1, \dots, K\}$ we have that $\left( \bm \theta_{G_k}^\ast, \bm \theta_{D_k}^\ast \right)$ is a solution to~(\ref{eq:hybrid_lossL0}) if and only if $\left( \bm \theta_{G_k}^\ast, \bm \theta_{D_k}^\ast \right)$ is a solution to~(\ref{eq:ensemble_loss}).
    \item Suppose that $t = 0$. Then we have that $\left( \bm \theta_{G}^\ast, \bm \theta_{D}^\ast \right)$ is a solution to~(\ref{eq:hybrid_lossL0}) for each $k \in \{ 1, \dots, K\}$ if and only if $\left( \bm \theta_{G}^\ast, \bm \theta_{D}^\ast \right)$ is a solution to~(\ref{eq:basic_loss}).
\end{enumerate}
\label{thm:L0Behavior}
\end{theorem}

Informally, \cref{thm:L0Behavior} shows that if we have $t=0$, then we essentially have a single GAN because all of the networks in the ensemble have the same parameters. If $t$ is large then we have an unconstrained problem such that the ensemble resembles the one in \cref{eq:ensemble_loss}. Therefore, this hybrid optimization problem trades off the parameter sharing between ensemble components in a way that allows us to compare performance of single GANs with ensembles.

Unfortunately, \cref{eq:hybrid_lossL0} is a combinatorial optimization problem and is computationally intractable. Experimentally, we relax~\cref{eq:hybrid_lossL0} to the following
\begin{align}
    \min_{ \bm \theta_{G_1}, \dots, \bm \theta_{G_K}}
    \left(
    \max_{\bm \theta_{D_1}, \dots, \bm \theta_{D_K}} 
    \sum_{k=1}^K V(\bm \theta_{G_k}, \bm \theta_{D_k}) 
     - \lambda \sum_{\substack{k=1 \\ j = k}}^K \|\bm \theta_{D_j} - \bm \theta_{D_k} \|_1 
    \right)
    + \lambda \sum_{\substack{k=1 \\ j = k}}^K \|\bm \theta_{G_j} - \bm \theta_{G_k} \|_1
    \label{eq:hybrid_lossL1}
\end{align}
in order to promote parameter sharing and have an almost everywhere differentiable regularization term that we can backpropagate through while training. 
Although~\cref{eq:hybrid_lossL1} is a relaxation of~\cref{eq:hybrid_lossL0}, we still have the same asymptotic behavior when we vary $\lambda$ as when we vary $t$ as shown in the supplementary material.

\begin{theorem}[Informal]
    \label{thm:L1BehaviorInformal}
    The optimization problem in (\ref{eq:hybrid_lossL0}) has the same asymptotic behavior as the optimization problem in (\ref{eq:hybrid_lossL1}).
\end{theorem}

\subsection{Optimality of ensembles}
\label{sec:ensemblesAreOptimal}

This next theorem shows that if we are able to learn each component's distribution, $P_{\mcX_k}$, then an ensemble can learn the whole data distribution $P_\mcX$.
\begin{theorem}
Suppose that 
$G_k^\ast$ is the network that generates $\mcX_k$ for each $k \in \{ 1, \dots, K\}$, i.e. $P_{G_k^\ast} = P_{\mcX_k}$. Under~\cref{ass:disconnected}, we can learn each $G_k^\ast$
by solving~(\ref{eq:ensemble_loss}) with $V$ being the objective function in~\cref{eq:basic_loss}.
\label{thm:optimality}
\end{theorem}
We know from~\cite{goodfellow2014generative} that a globally optimal solution is achieved when the distribution of the generated images equals $P_\mcX$. Hence, this theorem has an important consequence: Training an ensemble of networks is optimal under our current data model. 
Note that \cref{thm:optimality} also holds for other generative networks, such as variational autoencoders.

It is important to note that the condition ``$G_k$ is the network that generates $\mcX_k$'' is necessary but not too strong because we may have a distribution that cannot be learned by a generative network or that our network does not have enough capacity to learn. We do not care about such cases however, because we are studying the behavior of generative networks under \cref{ass:disconnected}.

\subsection{Relation to other GAN architectures}
\label{sec:GANsThatAreSecretlyEnsembles}

\paragraph{Relation to  cGANs.}~
We compare a cGAN to an ensemble of GANs. Recall from \cref{sec:GANsTreatingClassesDifferent} that a cGAN has parameters $\bm \theta_G$ and $\bm \theta_D$ that do not change with different labels but there are matrices $\mB_G$ and $\mB_D$ that do depend on the labels. Then, cGANs solve the following optimization problem:
\begin{align}
    \min_{ \bm \theta_{G_1}, \dots, \bm \theta_{G_K}}
    \max_{\bm \theta_{D_1}, \dots, \bm \theta_{D_K}} 
    \sum_{k=1}^K 
    V\left(\begin{bmatrix}\bm \theta_{G} \\ (\mB_G)_{\cdot, k} \end{bmatrix}, 
    \begin{bmatrix}\bm \theta_{D} \\ (\mB_D)_{\cdot, k} \end{bmatrix}
    \right) 
\end{align}
for each $k \in \{ 1, \dots, K\}$. Therefore, cGANs are an ensemble of GANs. Specifically, they solve the optimization problem \cref{thm:L0Behavior} with the additional constraint that the only parameters that can be different are the biases in the first layer. For cGANs which have multiple injections of the one-hot class vector at different layers, a similar result applies.

\begin{theorem}
\label{thm:cgans}
A cGAN is equivalent to an ensemble of GANs with parameter sharing among all parameters except for the biases in the first layer.
\end{theorem}

\paragraph{Relation to GM-GANs}
Another generative network that is related to ensembles is the GM-GAN. The first layer in GM-GANs transforms the latent distribution from isotropic Gaussian into a ensemble of Gaussians via the re-parameterization trick as discussed in \cref{sec:GANsTreatingClassesDifferent}. This new layer plays a similar role as the $\mB_{\cdot, k}$ in the cGAN comparison above, meaning that GM-GANs solve the optimization problem (\ref{eq:hybrid_lossL0}) with the additional constraint that the only parameters that can be different are the parameters in the first layer.

\begin{theorem}
\label{thm:gm-gans}
A GM-GAN is equivalent to an ensemble of GANs with parameter sharing among all parameters except for the first layer.
\end{theorem}

\subsection{Experimental Results}

In this section, we study how ensembles of WGAN~\cite{arjovsky2017wasserstein} compare with a single WGAN and a conditional WGAN. We use code from the authors' official repository~\cite{WGAN_repo} to train the baseline model. 
We modified this code to implement our ensembles of GANs and cGAN. For evaluating performance, we use the FID score~\cite{FID, FID_repo}, average MSE to the training data~\cite{metz2016unrolled, lipton2017precise}, and precision/recall~\cite{PR, PR_repo}. More details about the experimental setup are discussed in the supplementary material.

\subsubsection{Ensembles perform better than single networks}
\label{sec:exp_full_ensemble}
We consider a basic ensemble of WGANs where we simply copy over the WGAN architecture $10$ times  and train each network on the corresponding class of CIFAR-10; we call this the ``full ensemble''.
We compare this ensemble to the baseline WGAN  trained on CIFAR-10.

\cref{fig:baseline_vs_equi_ensemble} shows that the full ensemble of WGANs performs better than the single WGAN. It is not immediately clear, however, whether this boost in performance is due to the functional difference of having an ensemble or if it is happening because the ensemble has more parameters. The ensemble has $10$ times more parameters than the single WGAN, so the comparison is hard to make. Thus, we consider constraining the ensemble so that it has fewer parameters than the single WGAN.

\subsubsection{Ensembles with fewer total parameters still performs better than a single network}
\label{sec:exp_equi_ensemble}

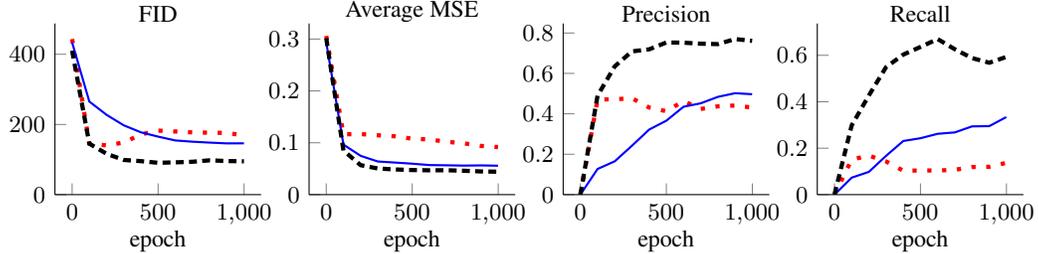
\begin{figure}[t]
\def\figScale{0.4}
\centering
\def\thickness{ultra thick}
\def\thicknessTwo{thick}

\def\colorOne{red}
\def\colorTwo{blue}
\def\colorThree{black}

\def\linestyle{densely dashed}
\def\linestyleTwo{loosely dotted}

    \setlength{\tabcolsep}{-0.14cm}
    \begin{tabular}{cccc}

    \begin{tikzpicture}
        \begin{axis}[ scale=\figScale,
            title={FID}, 
            xlabel={epoch},
            axis x line*=bottom,
            axis y line*=left,
            ymin=0,
            legend cell align={left},
            legend style={at={(1,1)},anchor=north east}]
            \addplot+[
                \colorOne,
                mark=none,
                \linestyleTwo,
                \thickness] 
            table [
                x index=0, 
                y index=1,
                col sep=comma] {csv/FID/WGAN_baseline.csv};
                
            \addplot+[
                \colorTwo,
                mark=none,
                \thicknessTwo] 
            table [
                x index=0, 
                y index=1,
                col sep=comma] {csv/FID/WGAN_equi_mixture.csv};
                
            \addplot+[
                \colorThree,
                mark=none,
                \linestyle,
                \thickness] 
            table [
                x index=0, 
                y index=1,
                col sep=comma] {csv/FID/WGAN_fullmixture.csv};
    
        \end{axis}
    \end{tikzpicture} 
    
    &
    \begin{tikzpicture}
        \begin{axis}[ scale=\figScale,
            title={Average MSE}, 
            xlabel={epoch},
            axis x line*=bottom,
            axis y line*=left,
            ymin=0,
            ymax=0.33,
            legend cell align={left},
            legend style={at={(1,1)},anchor=north east}]
            \addplot+[
                \colorOne,
                mark=none,
                \linestyleTwo,
                \thickness] 
            table [
                x index=0, 
                y index=1,
                col sep=comma] {csv/GOoF/WGAN_baseline.csv};
                
            \addplot+[
                \colorTwo,
                mark=none,
                \thicknessTwo] 
            table [
                x index=0, 
                y index=1,
                col sep=comma] {csv/GOoF/WGAN_equi_mixture.csv};
                
            \addplot+[
                \colorThree,
                mark=none,
                \linestyle,
                \thickness] 
            table [
                x index=0, 
                y index=1,
                col sep=comma] {csv/GOoF/WGAN_fullmixture.csv};
    
        \end{axis}
    \end{tikzpicture}
    
        &
        
    \begin{tikzpicture}
        \begin{axis}[ scale=\figScale,
            title={Precision}, 
            xlabel={epoch},
            axis x line*=bottom,
            axis y line*=left,
            ymin=0,
            legend cell align={left},
            legend style={at={(1,1)},anchor=north east}]
            \addplot+[
                \colorOne,
                mark=none,
                \linestyleTwo,
                \thickness] 
            table [
                x index=0, 
                y index=2,
                col sep=comma] {csv/PR/WGAN_baseline.csv};
                
            \addplot+[
                \colorTwo,
                mark=none,
                \thicknessTwo] 
            table [
                x index=0, 
                y index=2,
                col sep=comma] {csv/PR/WGAN_equi_mixture.csv};
                
            \addplot+[
                \colorThree,
                mark=none,
                \linestyle,
                \thickness] 
            table [
                x index=0, 
                y index=2,
                col sep=comma] {csv/PR/WGAN_fullmixture.csv};
    
        \end{axis}
    \end{tikzpicture} 
    
    &
    
    \begin{tikzpicture}
        \begin{axis}[ scale=\figScale,
            title={Recall}, 
            xlabel={epoch},
            axis x line*=bottom,
            axis y line*=left,
            ymin=0,
            legend cell align={left},
            legend style={at={(1,1)},anchor=north east}]
            \addplot+[
                \colorOne,
                mark=none,
                \linestyleTwo,
                \thickness] 
            table [
                x index=0, 
                y index=1,
                col sep=comma] {csv/PR/WGAN_baseline.csv};
                
            \addplot+[
                \colorTwo,
                mark=none,
                \thicknessTwo] 
            table [
                x index=0, 
                y index=1,
                col sep=comma] {csv/PR/WGAN_equi_mixture.csv};
                
            \addplot+[
                \colorThree,
                mark=none,
                \linestyle,
                \thickness] 
            table [
                x index=0, 
                y index=1,
                col sep=comma] {csv/PR/WGAN_fullmixture.csv};
    
        \end{axis}
    \end{tikzpicture} 
    \end{tabular}
            \vspace{-0.3cm}
        \caption{Ensembles of WGANs with fewer total parameters than a single WGAN perform better on CIFAR-10. 
        We do not have to sacrifice computation to achieve better performance, we just need models that capture the underlying structure of the data.
        The {\color{red} \bf dotted red} line is the baseline WGAN, the {\color{blue} \bf solid blue} line is the equivalent ensemble, and the {\color{black} \bf dashed black} line is the full ensemble. 
        }
    \label{fig:baseline_vs_equi_ensemble}
\end{figure}

The ``equivalent ensemble'' (3,120,040 total generator parameters) in \cref{fig:baseline_vs_equi_ensemble} still outperforms the single WGAN (3,476,704 generator parameters) showing that the performance increase comes from using the ensemble rather than just having larger capacity. 
In other words, considering ensembles of GANs allows for improved performance even though the ensemble is simpler than the original network in terms of number of parameters.

We see a performance boost as a result of increasing the number of parameters, in \cref{fig:baseline_vs_equi_ensemble}. Therefore, we perform better because of having a better model (an ensemble) as well as by having more parameters. Now, we investigate a way that we can further improve performance.
\subsubsection{Parameter sharing among ensemble components leads to better performance}
\label{sec:exp_L1_ensembles}

We study how the regularization penalty $\lambda$ affects performance.  As shown in \cref{thm:L1BehaviorInformal}, we can learn a model that is somewhere between an ensemble and a single network by using  $\ell^1$ regularization.

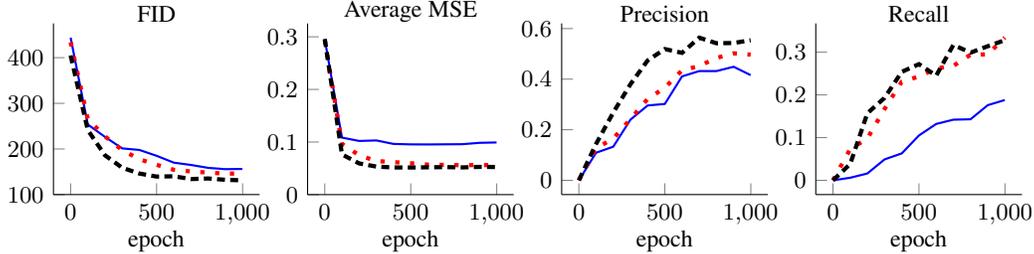
\begin{figure}

\def\thickness{ultra thick}
\def\thicknessTwo{thick}

\def\linestyle{densely dashed}
\def\linestyleTwo{loosely dotted}

\centering
    \def\figScale{0.4}
    \setlength{\tabcolsep}{-0.14cm}
    \centering
    \begin{tabular}{cccc}
    \begin{tikzpicture}
        \begin{axis}[ scale=\figScale,
            title={FID}, 
            xlabel={epoch},
            axis x line*=bottom,
            axis y line*=left,
            legend cell align={left},
            legend style={at={(1,1)}, anchor=north east}]
            
            \addplot+[
                blue,
                mark=none,
                \thicknessTwo] 
            table [
                x index=0, 
                y index=1,
                col sep=comma] {csv/FID/WGAN_equi_mixture_lambda01.csv};
            
            \addplot+[
                red,
                mark=none,
                \linestyleTwo,
                \thickness] 
            table [
                x index=0, 
                y index=1,
                col sep=comma] {csv/FID/WGAN_equi_mixture.csv};
                
            \addplot+[
                black,
                mark=none,
                \linestyle,
                \thickness] 
            table [
                x index=0, 
                y index=1,
                col sep=comma] {csv/FID/WGAN_equi_mixture_lambda001.csv};
    
        \end{axis}
    \end{tikzpicture}
    
    &
    
    \begin{tikzpicture}
        \begin{axis}[ scale=\figScale,
            title={Average MSE}, 
            xlabel={epoch},
            axis x line*=bottom,
            axis y line*=left,
            ymin=0,
            legend cell align={left},
            legend style={at={(1,1)}, anchor=north east}]
            
            \addplot+[
                blue,
                mark=none,
                \thicknessTwo] 
            table [
                x index=0, 
                y index=1,
                col sep=comma] {csv/GOoF/WGAN_equi_mixture_lambda01.csv};
            
            \addplot+[
                red,
                mark=none,
                \linestyleTwo,
                \thickness] 
            table [
                x index=0, 
                y index=1,
                col sep=comma] {csv/GOoF/WGAN_equi_mixture.csv};
                
            \addplot+[
                black,
                mark=none,
                \linestyle,
                \thickness] 
            table [
                x index=0, 
                y index=1,
                col sep=comma] {csv/GOoF/WGAN_equi_mixture_lambda001.csv};
    
        \end{axis}
    \end{tikzpicture}
    
    &
    
    \begin{tikzpicture}
        \begin{axis}[ scale=\figScale,
            title={Precision}, 
            xlabel={epoch},
            axis x line*=bottom,
            axis y line*=left,
            legend cell align={left},
            legend style={at={(1,1)}, anchor=north east}]
            
            \addplot+[
                blue,
                mark=none,
                \thicknessTwo] 
            table [
                x index=0, 
                y index=2,
                col sep=comma] {csv/PR/WGAN_equi_mixture_lambda01.csv};
            
            \addplot+[
                red,
                mark=none,
                \linestyleTwo,
                \thickness] 
            table [
                x index=0, 
                y index=2,
                col sep=comma] {csv/PR/WGAN_equi_mixture.csv};
                
            \addplot+[
                black,
                mark=none,
                \linestyle,
                \thickness] 
            table [
                x index=0, 
                y index=2,
                col sep=comma] {csv/PR/WGAN_equi_mixture_lambda001.csv};
    
        \end{axis}
    \end{tikzpicture}
    
    &
    
    \begin{tikzpicture}
        \begin{axis}[ scale=\figScale,
            title={Recall}, 
            xlabel={epoch},
            axis x line*=bottom,
            axis y line*=left,
            legend cell align={left},
            legend style={at={(1,1)}, anchor=north east}]
            
            \addplot+[
                blue,
                mark=none,
                \thicknessTwo] 
            table [
                x index=0, 
                y index=1,
                col sep=comma] {csv/PR/WGAN_equi_mixture_lambda01.csv};
            
            \addplot+[
                red,
                mark=none,
                \linestyleTwo,
                \thickness] 
            table [
                x index=0, 
                y index=1,
                col sep=comma] {csv/PR/WGAN_equi_mixture.csv};
                
            \addplot+[
                black,
                mark=none,
                \linestyle,
                \thickness] 
            table [
                x index=0, 
                y index=1,
                col sep=comma] {csv/PR/WGAN_equi_mixture_lambda001.csv};
    
        \end{axis}
    \end{tikzpicture}
    \end{tabular}
    
        \vspace{-0.3cm}
        \caption{
        This shows a sweet spot in terms of the performance measures when we regularize the optimization problem in expression (\ref{eq:hybrid_lossL1}) with different values of $\lambda$. Each curve is calculated using the equivalent ensemble of WGANs discussed in \cref{sec:exp_equi_ensemble}. We see that as we increase $\lambda$ to $0.001$, the performance increases but then decreases when we continue to increase $\lambda$ to $0.01$. This implies that there is an optimal value for $\lambda$ that can be found via hyperparameter tuning.
        The {\color{blue} \bf solid blue} line is the equivalent ensemble with $\lambda = 0.01$, the {\color{red} \bf dotted red} line is the equivalent ensemble WGAN, and the {\color{black} \bf dashed black} line is the equivalent ensemble with $\lambda = 0.001$.}
    \label{fig:L1_ensemble_sweet_spot}
\end{figure}

In \cref{fig:L1_ensemble_sweet_spot}, the performance increases when we increase $\lambda$ in the equivalent ensemble from $0$ to $0.001$, implying that there is some benefit to regularization. 
Recall that by having $\lambda > 0$, we force parameter sharing between generator and discriminator networks.  
This performance increase is likely data dependent and has to do with the structure of the underlying data $\mcX$. For example, we can have pictures of badgers $(\mcX_1)$ and zebras $(\mcX_2)$ in our dataset and they are disconnected. However, the backgrounds of these images are likely similar so that there is some benefit in $G_1$ and $G_2$ treating these images similarly, if only to remove the background.

As we increase $\lambda$ from $0.001$ to $0.01$ we notice that performance decreases. This means that there is a sweet spot and we may be able to find an optimal $0 < \lambda^\ast < 0.01$ via hyperparameter tuning. 
We know that the performance is not monotonic with respect to $\lambda$ because it decreases and then increases again; in other words, the performance has a minima that is not at $\lambda = 0$ or $\lambda \rightarrow \infty$.
The optimization problem in expression (\ref{eq:hybrid_lossL1}) therefore can be used to find a better ensemble than the  equivalent ensemble used in \cref{sec:exp_equi_ensemble} which still has fewer parameters than the baseline WGAN.

\subsubsection{Ensembles outperform cGANs}
\label{sec:cGANs_suck}

In this section, we modify a WGAN to be conditional and call it cWGAN. This cWGAN is trained on CIFAR-10, and we compare cWGAN to ensembles of WGANs. We do this because we showed
in \cref{sec:GANsThatAreSecretlyEnsembles} that cGANs are an ensemble of GANs that only have the bias in the first layer change between ensemble networks.

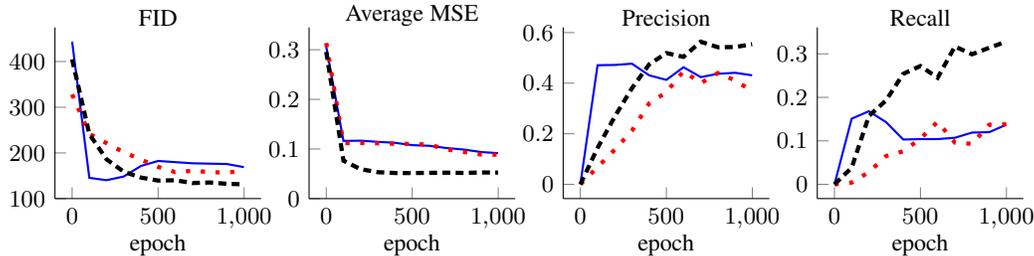
\begin{figure}
\def\thickness{ultra thick}
\def\thicknessTwo{thick}

\def\linestyle{densely dashed}
\def\linestyleTwo{loosely dotted}

    \def\figScale{0.4}
    \setlength{\tabcolsep}{-0.14cm}
    \centering
    \begin{tabular}{cccc}
    \begin{tikzpicture}
        \begin{axis}[ scale=\figScale,
            title={FID},
            xlabel={epoch},
            axis x line*=bottom,
            axis y line*=left,
            legend cell align={left},
            legend style={at={(1,1)}, anchor=north east}]
            
            \addplot+[
                blue,
                mark=none,
                \thicknessTwo] 
            table [
                x index=0, 
                y index=1,
                col sep=comma] {csv/FID/WGAN_baseline.csv};
            
            \addplot+[
                red,
                mark=none,
                \linestyleTwo,
                \thickness] 
            table [
                x index=0, 
                y index=1,
                col sep=comma] {csv/FID/WGAN_cgan.csv};
                
            \addplot+[
                black,
                mark=none,
                \linestyle,
                \thickness] 
            table [
                x index=0, 
                y index=1,
                col sep=comma] {csv/FID/WGAN_equi_mixture_lambda001.csv};

        \end{axis}
    \end{tikzpicture}
    
    &
    
    \begin{tikzpicture}
        \begin{axis}[ scale=\figScale,
            title={Average MSE}, 
            xlabel={epoch},
            axis x line*=bottom,
            axis y line*=left,
            ymin=0,
            legend cell align={left},
            legend style={at={(1,1)}, anchor=north east}]
            
            \addplot+[
                blue,
                mark=none,
                \thicknessTwo] 
            table [
                x index=0, 
                y index=1,
                col sep=comma] {csv/GOoF/WGAN_baseline.csv};
            
            \addplot+[
                red,
                mark=none,
                \linestyleTwo,
                \thickness] 
            table [
                x index=0, 
                y index=1,
                col sep=comma] {csv/GOoF/WGAN_cgan.csv};
                
            \addplot+[
                black,
                mark=none,
                \linestyle,
                \thickness] 
            table [
                x index=0, 
                y index=1,
                col sep=comma] {csv/GOoF/WGAN_equi_mixture_lambda001.csv};

        \end{axis}
    \end{tikzpicture}
    
    &
    
    \begin{tikzpicture}
        \begin{axis}[ scale=\figScale,
            title={Precision}, 
            xlabel={epoch},
            axis x line*=bottom,
            axis y line*=left,
            legend cell align={left},
            legend style={at={(1,1)}, anchor=north east}]
            
            \addplot+[
                blue,
                mark=none,
                \thicknessTwo] 
            table [
                x index=0, 
                y index=2,
                col sep=comma] {csv/PR/WGAN_baseline.csv};
            
            \addplot+[
                red,
                mark=none,
                \linestyleTwo,
                \thickness] 
            table [
                x index=0, 
                y index=2,
                col sep=comma] {csv/PR/WGAN_cgan.csv};
                
            \addplot+[
                black,
                mark=none,
                \linestyle,
                \thickness] 
            table [
                x index=0, 
                y index=2,
                col sep=comma] {csv/PR/WGAN_equi_mixture_lambda001.csv};

        \end{axis}
    \end{tikzpicture}
    
    &
    
    \begin{tikzpicture}
        \begin{axis}[ scale=\figScale,
            title={Recall}, 
            xlabel={epoch},
            axis x line*=bottom,
            axis y line*=left,
            legend cell align={left},
            legend style={at={(1,1)}, anchor=north east}]
            
            \addplot+[
                blue,
                mark=none,
                \thicknessTwo] 
            table [
                x index=0, 
                y index=1,
                col sep=comma] {csv/PR/WGAN_baseline.csv};
            
            \addplot+[
                red,
                mark=none,
                \linestyleTwo,
                \thickness] 
            table [
                x index=0, 
                y index=1,
                col sep=comma] {csv/PR/WGAN_cgan.csv};
                
            \addplot+[
                black,
                mark=none,
                \linestyle,
                \thickness] 
            table [
                x index=0, 
                y index=1,
                col sep=comma] {csv/PR/WGAN_equi_mixture_lambda001.csv};

        \end{axis}
    \end{tikzpicture}
    \end{tabular}
            \vspace{-0.3cm}

    \caption{This shows that although cGANs are a type of ensemble, they do not perform as well as regularized ensembles using the optimization in (\ref{eq:hybrid_lossL1}). Here, cWGAN actually performs similarly to the baseline WGAN even though it takes into consideration class information.
    The {\color{blue} \bf solid blue} line is the baseline, the {\color{red} \bf dotted red} line is the cWGAN, and the {\color{black} \bf dashed black} line is the equivalent ensemble with $\lambda = 0.001$.
    }
    \label{fig:cgans_suck}
\end{figure}

As can be seen from \cref{fig:cgans_suck}, ensembles perform better than the cWGAN. The baseline WGAN model actually performs similarly to the cWGAN, which implies that the conditioning is not helping in this specific case. We hypothesize that our model ($\lambda = 0.001$) performs better because there are more parameters that are free in the optimization, instead of just the bias in the first layer. Thus, although cGANs are widely used, an ensemble with the regularization described in \cref{sec:ensemblesIntro} can outperform them because the ensemble captures the disconnected structure of the data better.

\section*{Acknowledgements}
This material is based upon work supported by the National Science Foundation under Grant No. 1842494. Any opinions, findings, and conclusions or recommendations expressed in this material are those of the author(s) and do not necessarily reflect the views of the National Science Foundation.

\bibliography{ref}
\bibliographystyle{plainLL}

\end{document}


\maketitle

\section{Proofs}
We first show that under \AssDisconnected{}, our data is disconnected.

\showTheorem{
\begin{lemma}
    $\mcX$ is a disconnected set.
    \label{lemma:disconnected}
\end{lemma}
}{}

\begin{proof}[\bf Proof of~\cref{lemma:disconnected}]
    From \AssDisconnected{} take $A = \mcX_1$ and $B = \bigcupdot_{k=2}^K \mcX_k$. Let $d = \inf \{ |\vx - \vy| : \vx \in A, \vy \in B \}$, where $|\cdot|$ is the standard Euclidean metric. Notice that $d > 0$ otherwise, $A$ and $B$ share a limit point which is impossible since they are both closed in $\bbR^p$. Hence, we can form an open cover of $A$ and $B$ as follows. Let $A' = \bigcup_{\vx \in A} \{ \vy \in \mcX : |\vy - \vx| < \frac{d}{4} \}$ which is a union of open balls (with respect to the subset topology on $\mcX$) centered at $\vx$ with radius $\frac{d}{4}$ for each $\vx \in A$. Notice that $A'$ is an open set because it is a union of open sets. Moreover, it is easy to see that $A' = A$ because the radius of the balls is smaller than the smallest distance between $B$ and $A$. We construct an open cover of $B$ the same way to show that $B$ is open. Thus, we have that $\mcX$ is disconnected. 
\end{proof}

\begin{remark}
Note that if our data is disconnected, it doesn't necessarily follow \AssDisconnected{}. This means that \AssDisconnected{} is a stronger condition than just having disconnected data. This can be seen by the following counter-example. We denote a truncated Gaussian as $\mcN(\mu,\sigma^2)|_S$, where the distribution is non-zero on $S$. Let
\[
    P_{\mcX} 
    = \frac{1}{2} P_{\mcX_1} + \frac{1}{2} P_{\mcX_2} 
    = \frac{1}{2} \mcN(0,1)|_{(-\infty,0)} + \frac{1}{2} \mcN(0,1)|_{(0,\infty)}
\]
be the true distributions. Note that $(-\infty,0)$ and $(0,\infty)$ are disconnected but do not follow \AssDisconnected{} because they are open sets. Moreover, we can learn this distribution easily by letting $G$ be the identity function and having $P_z = \mcN(0,1)$; it is trivial to show that this results in $P_G = P_\mcX$. Hence, disconnected data is too weak of an assumption---we need there to be a non-zero distance between our disconnected sets and that is what \AssDisconnected{} captures.
\end{remark}

\showTheorem{
\begin{theorem}
The probability of generating samples outside of $\mcX$ is positive:
$
    P_G (\vx \in \bbR^p \backslash \mcX) > 0.
$
\label{thm:probPositive}
\end{theorem}
}{}

\begin{proof}[\bf Proof of~\cref{thm:probPositive}]
We define $B = G^{-1}(\mcX)$. Since, $\mcX$ is disconnected and closed in $\bbR^p$, we have that $B$ is disconnected and closed in $\bbR^\ell$ because $G$ is continuous (Theorem 4.8 and 4.22 in~\cite{babyRudin}). Since $B$ is closed in $\bbR^\ell$, this means that $\bbR^\ell \backslash B$ is an open set. Moreover, $\bbR^\ell \backslash B$ is not empty because we know that $\bbR^\ell$ is connected and $B$ is not. We also know that the Lebesgue measure $\lambda$ of a nonempty, open set is positive, hence we have that 
\[
    \lambda(\bbR^\ell \backslash B) > 0.
\]
Since $\lambda$ is equivalent to $P_\vz$, we have that $P_\vz(\vz \in \bbR^\ell \backslash B) > 0$. Thus, 
\[
    P_G(\vx \in \bbR^p \backslash \mcX)
    = P_\vz(\vz \in G^{-1}(\bbR^p \backslash \mcX))
    = P_\vz(\vz \in \bbR^\ell \backslash  B)
    > 0,
\]
as desired.
\end{proof}

\showTheorem{
\begin{corollary}
Under \AssDisconnected{}, $d(P_G, P_\mcX) > 0$ for any distance metric $d$ and any learned distribution $P_G$. 
\label{cor:distPositive}
\end{corollary}
}{}
\begin{proof}[\bf Proof of~\cref{cor:distPositive}]
    Since the data lies only on $\mcX$, we know that $P_\mcX(\vx_\text{data} \in \bbR \backslash \mcX) = 0$ for any valid probability measure. However, we have that $P_G(\vx \in \bbR^p \backslash \mcX) > 0$. Hence, $d(P_{G}, P_\mcX) > 0$ for any metric $d$.
\end{proof}

\showTheorem{
\begin{theorem}[\bf Truncating the latent space reduces error]
\label{thm:truncatingPz}
Suppose that $P_\vz(Z) > 0$ and let the generator $G$ learn a proportionally correct distribution over $\mcX$. In other words, there exists a real number $c \in \bbR$ so that
\[
    P_G(A) = c P_\mcX(A) \quad \quad A \in \mcF', A \subset \mcX.
\]
Then, we use the truncated latent distribution defined by $P_{\vz_T}(B) = 0$ for all $B \in \mcF$ that satisfy $B \cap Z = \emptyset$. This allows us to learn the data distribution exactly, i.e.
\[
    P_{G|_Z}(A) = P_\mcX(A) \quad \quad A \in \mcF'.
\]
\end{theorem}
}{}
\begin{proof}[\bf Proof of~\cref{thm:truncatingPz}]
    The truncated latent distribution is denoted $P_{\vz_T}$ and is defined as
    \[
        P_{\vz_T}(B) = \frac{P_\vz(B \cap Z)}{P_\vz(Z)}
    \]
    for any set $B \in \mcF$. Hence, we have that
    \begin{align*}
        P_{G|_Z}(A) 
        &= P_{\vz_T}( G^{-1}(A) ) 
        \\
        &= \frac{P_\vz(G^{-1}(A) \cap Z)}{P_\vz(Z)}
        \\
        &= \frac{1}{P_\vz(Z)}P_\vz(G^{-1}(A) \cap G^{-1}(\mcX))
        \\
        &= \frac{1}{P_\vz(Z)}P_\vz(G^{-1}(A \cap \mcX))
        \\
        &= \frac{1}{P_\vz(Z)}P_G(A \cap \mcX)
        \\
        &= \frac{c}{P_\vz(Z)}P_\mcX(A \cap \mcX)
        \\
        &= \frac{c}{P_\vz(Z)}P_\mcX(A)
    \end{align*}
    for any $A \in \mcF'$. The last equality is true because $P_\mcX(\mcX) = 1$, so that any points outside of $\mcX$ have zero probability. For the result above, set $A = \bbR^p$ to see that $c = P_\vz(Z)$, implying that
    \[
        P_{G|_Z}(A) = P_\mcX(A) \quad \quad A \in \mcF',
    \]
    as desired.
\end{proof}

\showTheorem{
\begin{theorem}
Let $G$ and $D$ be the generator and discriminator network in a GAN. Suppose that for $k \in \{ 1, \dots, K\}$ we have that $G_k$ and $D_k$ have the same architectures as $G$ and $D$, respectively. Then,
\begin{enumerate}[label=\roman*)]
    \item Suppose that $t \geq \max\Big\{ K\frac{K-1}{2}|\bm \theta_D|, K\frac{K-1}{2}|\bm \theta_G|\Big\}$. Then for all $k \in \{ 1, \dots, K\}$ we have that $\left( \bm \theta_{G_k}^\ast, \bm \theta_{D_k}^\ast \right)$ is a solution to~\EqHybridLossLZero{} if and only if $\left( \bm \theta_{G_k}^\ast, \bm \theta_{D_k}^\ast \right)$ is a solution to~\EqEnsembleLoss{}.
    \item Suppose that $t = 0$. Then we have that $\left( \bm \theta_{G}^\ast, \bm \theta_{D}^\ast \right)$ is a solution to~\EqHybridLossLZero{} for each $k \in \{ 1, \dots, K\}$ if and only if $\left( \bm \theta_{G}^\ast, \bm \theta_{D}^\ast \right)$ is a solution to~\EqBasicLoss{}.
\end{enumerate}
\label{thm:L0Behavior}
\end{theorem}
}{}
\begin{proof}[\bf Proof of~\cref{thm:L0Behavior}]
    First we prove i). If $t \geq \max\Big\{ K\frac{K-1}{2}|\bm \theta_D|, K\frac{K-1}{2}|\bm \theta_G|\Big\}$ then the constraints on~\EqHybridLossLZero{} are unnecessary so that the problem reduces to 
    \begin{align*}
        \min_{ \bm \theta_{G_1}, \dots, \bm \theta_{G_K}}
        \max_{\bm \theta_{D_1}, \dots, \bm \theta_{D_K}} 
        \sum_{k=1}^K V(\bm \theta_{G_k}, \bm \theta_{D_k})
        &=\min_{ \bm \theta_{G_1}, \dots, \bm \theta_{G_K}}
        \sum_{k=1}^K \max_{\bm \theta_{D_k}} 
        V(\bm \theta_{G_k}, \bm \theta_{D_k})
        \\
        &=\sum_{k=1}^K \min_{ \bm \theta_{G_k}}
        \max_{\bm \theta_{D_k}} 
        V(\bm \theta_{G_k}, \bm \theta_{D_k}),
    \end{align*}
    which is equivalent to solving the optimization problem
    \[
        \min_{ \bm \theta_{G_k}}
        \max_{\bm \theta_{D_k}} 
        V(\bm \theta_{G_k}, \bm \theta_{D_k}), \quad k \in \{ 1, \dots, K\}.
    \]
    Thus, i) is shown.

    Next, we prove ii). Suppose that $t = 0$ and that $\left( \bm \theta_{G}^\ast, \bm \theta_{D}^\ast \right)$ is a solution to~\EqHybridLossLZero{} for each $k \in \{ 1, \dots, K\}$. This means that 
        \begin{align*}
        &\min_{ \bm \theta_{G_1}, \dots, \bm \theta_{G_K}}
        \left(
        \max_{\bm \theta_{D_1}, \dots, \bm \theta_{D_K}} 
        \sum_{k=1}^K V(\bm \theta_{G_k}, \bm \theta_{D_k}) 
        \text{ s.t.} \sum_{\substack{k=1 \\ j = k}}^K \|\bm \theta_{D_j} - \bm \theta_{D_k} \|_0 = 0
        \right)
        \text{ s.t.} \sum_{\substack{k=1 \\ j = k}}^K \|\bm \theta_{G_j} - \bm \theta_{G_k} \|_0 = 0
        \\
        &=
        \min_{ \bm \theta_{G}}
        \max_{\bm \theta_{D}} 
        \sum_{k=1}^K V(\bm \theta_{G}, \bm \theta_{D})
        \\
        &=
        \min_{ \bm \theta_{G}}
        \max_{\bm \theta_{D}} 
        K V(\bm \theta_{G}, \bm \theta_{D}),
    \end{align*}
    which is equivalent to the optimization problem in~\EqBasicLoss{}, as desired.
\end{proof}

\showTheorem{
\begin{theorem}
Let $G$ and $D$ be the generator and discriminator network in a GAN. Suppose that for $k \in \{ 1, \dots, K\}$ we have that $G_k$ and $D_k$ have the same architectures as $G$ and $D$, respectively. Then,
\begin{enumerate}[label=\roman*)]
    \item Suppose that $\lambda = 0$. Then for all $k \in \{ 1, \dots, K\}$ we have that $\left( \bm \theta_{G_k}^\ast, \bm \theta_{D_k}^\ast \right)$ is a solution to~\EqHybridLossLOne{} if and only if $\left( \bm \theta_{G_k}^\ast, \bm \theta_{D_k}^\ast \right)$ is a solution to~\EqEnsembleLoss{}.
    \item Suppose that $\lambda \rightarrow \infty$. Then we have that $\left( \bm \theta_{G}^\ast, \bm \theta_{D}^\ast \right)$ is a solution to~\EqHybridLossLOne{} for each $k \in \{ 1, \dots, K\}$ if and only if $\left( \bm \theta_{G}^\ast, \bm \theta_{D}^\ast \right)$ is a solution to~\EqBasicLoss{}.
\end{enumerate}
\label{thm:L1Behavior}
\end{theorem}
}{}

\begin{proof}[\bf Proof of~\cref{thm:L1Behavior}]
    First we prove i). If $\lambda = 0$ then we have that
    \[
        \lambda \sum_{\substack{k=1 \\ j = k}}^K \|\bm \theta_{D_j} - \bm \theta_{D_k} \|_1 
        = 
        \lambda \sum_{\substack{k=1 \\ j = k}}^K \|\bm \theta_{G_j} - \bm \theta_{G_k} \|_1
        =0
    \]
    on~\EqHybridLossLOne{}. Hence, the problem reduces to the unconstrained problem of~\EqEnsembleLoss{}.
    
    Next, we prove ii). Since $\lambda \rightarrow \infty$, any solution where $\bm \theta_{D_k} \ne \bm \theta_{D_j}$ or $\bm \theta_{G_k} \ne \bm \theta_{G_j}$ for all $j, k \in \{ 1, \dots, K\}$ is suboptimal. Consequently, it means that the optimization problem in~\EqHybridLossLOne{} reduces to
    \begin{align*}
        &\min_{ \bm \theta_{G_1}, \dots, \bm \theta_{G_K}}
        \left(
        \max_{\bm \theta_{D_1}, \dots, \bm \theta_{D_K}} 
        \sum_{k=1}^K V(\bm \theta_{G_k}, \bm \theta_{D_k}) 
        - \lambda \sum_{\substack{k=1 \\ j = k}}^K \|\bm \theta_{D_j} - \bm \theta_{D_k} \|_1 
        \right)
        + \lambda \sum_{\substack{k=1 \\ j = k}}^K \|\bm \theta_{G_j} - \bm \theta_{G_k} \|_1
        \\
        &=
        \min_{ \bm \theta_G}
        \max_{\bm \theta_D} 
        \sum_{k=1}^K V(\bm \theta_{G}, \bm \theta_{D}) 
        \\
        &=
        \min_{ \bm \theta_{G}}
        \max_{\bm \theta_{D}} 
        K V(\bm \theta_{G}, \bm \theta_{D}),
    \end{align*}
    which is equivalent to the optimization problem in~\EqBasicLoss{}, as desired.
\end{proof}

\showTheorem{
\begin{theorem}
Suppose that 
$G_k^\ast$ is the network that generates $\mcX_k$ for each $k \in \{ 1, \dots, K\}$, i.e. $P_{G_k^\ast} = P_{\mcX_k}$. 
Under \AssDisconnected{}, we can learn each $G_k^\ast$
by solving~\EqEnsembleLoss{} with $V$ being the objective function in~\EqBasicLoss{}.
\label{thm:optimality}
\end{theorem}
}{}
\begin{proof}[\bf Proof of~\cref{thm:optimality}]
    Note that $P_\mcX$ is the total data distribution and that $P_{\mcX_k}$ is the distribution of each disconnected set. This means that
    \[
        P_\mcX = \sum_{k=1}^K \pi_k P_{\mcX_k}
    \]
    for some mixture coefficients $\alpha_k > 0$ so that $\sum_{k=1}^K \alpha_k = 1$.

    Fix an arbitrary $k \in \{ 1, \dots, K\}$. Since $P_{G_k^\ast} = P_{\mcX_k}$, we have that
    \[
        \min_{\bm \theta_{G_k}}\max_{\bm \theta_{D_k}} V(\bm \theta_{G = _k}, \bm \theta_{D_k})
        =
        \min_{\bm \theta_{G_k}} \max_{\bm \theta_{D_k}} \bbE_{\vx \sim P_{\mcX_k}}
        [\log D_k( \vx)] 
        + 
        \bbE_{\vz \sim P_\vz} [\log (1 - D_k(G_k(\vz)))]
    \]
    has a solution of $P_{G_k^\ast} = P_{\mcX_k}$ from Theorem 1 of~\cite{goodfellow2014generative}. Since this is true for every $k$ and since $P_\mcX = \sum_{k=1}^K \pi_k P_{\mcX_k}$, we learn the complete data distribution.
\end{proof}

\showTheorem{
\begin{theorem}
\label{thm:cgans}
A cGAN is equivalent to an ensemble of GANs with parameter sharing among all parameters except for the biases in the first layer.
\end{theorem}
}{}
\begin{proof}[\bf Proof of~\cref{thm:cgans}]
    We begin with the generic optimization problem from~\EqHybridLossLZero{}:
    \begin{align*}
        &
        \min_{ \bm \theta_{G_1}, \dots, \bm \theta_{G_K}}
        \left(
        \max_{\bm \theta_{D_1}, \dots, \bm \theta_{D_K}} 
        \sum_{k=1}^K V(\bm \theta_{G_k}, \bm \theta_{D_k}) 
        \text{ s.t.} \sum_{\substack{k=1 \\ j = k}}^K \|\bm \theta_{D_j} - \bm \theta_{D_k} \|_0 \leq t
        \right)
        \text{ s.t.} \sum_{\substack{k=1 \\ j = k}}^K \|\bm \theta_{G_j} - \bm \theta_{G_k} \|_0 \leq t
        \\
        &=
        \min_{ \bm \theta_{G_1}, \dots, \bm \theta_{G_K}}
        \left(
        \max_{\bm \theta_{D_1}, \dots, \bm \theta_{D_K}} 
        \sum_{k=1}^K V(\bm \theta_{G_k}, \bm \theta_{D_k}) 
        \text{ s.t. } C_D
        \right)
        \text{ s.t. } C_G
        \\
        &=
        \min_{ \bm \theta_{G_1}, \dots, \bm \theta_{G_K}}
        \left(
        \max_{\bm \theta_{D_1}, \dots, \bm \theta_{D_K}} 
        \sum_{k=1}^K 
        V\left(\begin{bmatrix}\bm \theta_{G_k}' \\ (\mB_G)_{\cdot, k} \end{bmatrix}, 
        \begin{bmatrix}\bm \theta_{D_k}' \\ (\mB_D)_{\cdot, k} \end{bmatrix}
        \right) 
        \text{ s.t. } C_D
        \right)
        \text{ s.t. } C_G
    \end{align*}
    where we simply use the name $C_D$ for the constraint $\sum_{\substack{k=1 \\ j = k}}^K \|\bm \theta_{D_j} - \bm \theta_{D_k} \|_0 \leq t$ and similarly for $C_G$. This is purely for notational convenience. Likewise, we simply denote $\bm \theta_{G_k}$ as $[ (\bm \theta_{G_k}')^T \  (\mB_G)_{\cdot, k}^T ]^T$ and similarly for $\bm \theta_{D_k}$, for each $k$. Keep in mind that $\mB_G$ and $\mB_D$ are matrices such that the $k$-th column is the the bias of the first layer of the $k$-th network in the ensemble.
    So far, we have only introduced notational changes.
    
    Consider what happens if we change the constraints to
    \begin{align*}
        C_D' &= \sum_{\substack{k=1 \\ j = k}}^K \|\bm \theta_{D_j}' - \bm \theta_{D_k}' \|_0 = 0
        \\
        C_G' &= \sum_{\substack{k=1 \\ j = k}}^K \|\bm \theta_{G_j}' - \bm \theta_{G_k}' \|_0 = 0.
    \end{align*}
    We have that $\mB_G$ and $\mB_D$ are unconstrained and that $\bm \theta_{G_k}'$ is forced to be equal to $\bm \theta_{G_j}'$ for all $k$ and $j$. Similarly $\bm \theta_{D_k}' = \bm \theta_{D_j}'$ for all $k$ and $j$. Hence, we can say that the optimization problem above with the new constraint is
    \begin{align*}
        &\min_{ \bm \theta_{G_1}, \dots, \bm \theta_{G_K}}
        \left(
        \max_{\bm \theta_{D_1}, \dots, \bm \theta_{D_K}} 
        \sum_{k=1}^K 
        V\left(\begin{bmatrix}\bm \theta_{G_k}' \\ (\mB_G)_{\cdot, k} \end{bmatrix}, 
        \begin{bmatrix}\bm \theta_{D_k}' \\ (\mB_D)_{\cdot, k} \end{bmatrix}
        \right) 
        \text{ s.t. } C_D'
        \right)
        \text{ s.t. } C_G'
        \\
        &=
        \min_{ \bm \theta_{G_1}, \dots, \bm \theta_{G_K}}
        \max_{\bm \theta_{D_1}, \dots, \bm \theta_{D_K}} 
        \sum_{k=1}^K 
        V\left(\begin{bmatrix}\bm \theta_{G} \\ (\mB_G)_{\cdot, k} \end{bmatrix}, 
        \begin{bmatrix}\bm \theta_{D} \\ (\mB_D)_{\cdot, k} \end{bmatrix}
        \right),
    \end{align*}
    which is equivalent to the cGAN optimization problem. Here, we just define $\bm \theta_G$ to be shorthand for any one of the $\bm \theta_{G_k}$ vectors, since they are all the same.
    
    Hence, a cGAN is equivalent to solving the ensemble optimization problem in~\EqHybridLossLZero{} with a modified constraint.
    
\end{proof}

\showTheorem{
\begin{theorem}
\label{thm:gm-gans}
A GM-GAN is equivalent to an ensemble of GANs with parameter sharing among all parameters except for the first layer.
\end{theorem}
}{}
\begin{proof}[\bf Proof of~\cref{thm:gm-gans}]
The proof for this is very similar to the proof for \cref{thm:cgans}.
\end{proof}

\section{Estimation of ensemble parameters}
\label{sec:estimatingParameters}
In \SecEnsemblesIntro{}, we assume that $k \sim p_k$ is a multinomial distribution of degree $K$ parameters: $\pi_i$ for $i=1, \dots, K$. Using the maximum likelihood estimator~\cite{bishop2006pattern} we obtain 
\[
    \hat\pi^{\text{MLE}}_i = \frac{1}{N}\sum_{j=1}^N \mathds{1}(\vy_j = i)
\]
for $i = \{1, \dots, K\}$. For datasets like MNIST~\cite{mnist} and CIFAR-10~\cite{cifar10}, $k$ is a uniformly distributed random variable. For others one may have to calculate $p_k$ based on class imbalances.

\section{Experimental details}
In this section we describe the details of our experiments.
\subsection{Performance measures}
We use FID~\cite{FID}, average MSE~\cite{metz2016unrolled}, precision, and recall~\cite{PR} to evaluate our models. 

For FID, precision, and recall we use the official repositories~\cite{FID_repo,PR_repo}. For each of these methods, we compare a set of generated images to a set of images from the training set. For the FID calculation, we use the precalculated statistics for CIFAR-10 and compare to 10,000 generated images from our trained networks. For precision and recall, we compare 10,000 generated images to 10,000 images in the training set. All other parameters are left the same.

For the average MSE calculation, we use the algorithm introduced in~\cite{lipton2017precise}, which was empirically shown to work 100\% of the time on DCGAN architecture, such as WGAN. We modified the  code in~\cite{GOoF_repo} so that it can be run with multiple restarts if desired.
We ran our experiments with 1000 iterations and 5 restarts. We ran the code on 100 training images.

\subsection{Baseline model}
For the baseline model, we ran the default WGAN code for 1000 epochs on CIFAR-10. All other parameters are left at their default values.

\subsection{Full ensemble}
To create the full ensemble, we just copied over the baseline model $10$ times and trained each network pair $(G_k, D_k)$ in the ensemble on a single class of CIFAR-10. The training also lasted for 1000 epochs. This is equivalent to solving the optimization problem in \EqEnsembleLoss{}.

\subsection{Equivalent ensemble}

Normally WGAN is trained with the following two architecture parameters: ngf = $64$ and ndf = $64$. However, to get 10\% of the parameters we trained each ensemble component with ngf = $15$ and ndf = $20$. The depth of the generator and discriminator in the equivalent ensemble are the same as in the single WGAN, however, we modify the width of each corresponding layer so that the total parameters are fewer in the ensemble than in the single WGAN. Specifically, the generator of the WGAN has $3,576,704$ parameters and each generator of the equivalent ensemble has $312,004$ parameters. The discriminator of the WGAN has $2,765,568$ parameters and each discriminator of the equivalent ensemble has $272,880$ parameters. Reducing the width of each layer is not necessarily the optimal way to reducing parameters in a network. We do this because it is easy and effective, not because we are trying to reduce parameters in an optimal way, which is out of the scope of this paper. This is equivalent to solving the optimization problem in \EqEnsembleLoss{}.

\subsection{Regularized ensembles}
For all the ensembles with $\lambda > 0$, we use the equivalent ensemble architecture, while solving \EqHybridLossLOne{}.

\subsection{The cGAN model}
For this architecture, we modify the baseline architecture and concatenate the class label, represented as a one-hot vector, to the input of the generator and discriminator networks.

\bibliography{ref}
\bibliographystyle{plainLL}